\documentclass[twocolumn]{article}

\usepackage[utf8]{inputenc}

\usepackage[backend=bibtex]{biblatex}

\usepackage{comment}

\usepackage{graphicx}

\usepackage{algorithm}

\usepackage{placeins}

\usepackage{amsfonts}

\usepackage{algpseudocode}

\usepackage{amsmath}

\usepackage{amsthm}

\usepackage{newfloat}

\usepackage{listings}

\graphicspath{{./figures/}}

\title{Plan-Based Derivation of General Functional Structures in Product Design}

\author{Philipp Rosenthal, Niels Demke, Frank Mantwill, Oliver Niggemann}

\date{November 30, 2022}

\begin{document}

\maketitle

\begin{abstract}

In product design, a decomposition of the overall product function into a set of smaller, interacting functions is usually considered a crucial first step for any computer-supported design tool. Here, we propose a new approach for the decomposition of functions especially suited for later solutions based on Artificial Intelligence.

The presented approach defines the decomposition problem in terms of a planning problem---a well established field in Artificial Intelligence. For the planning problem, logic-based solvers can be used to find solutions that compute a useful function structure for the design process. Well-known function libraries from engineering are used as atomic planning steps. 

The algorithms are evaluated using two different application examples to ensure the transferability of a general function decomposition.
\end{abstract}

\section{Introduction}


Feldman et al.~\cite{Feldman.2019} consider the design domain as the next frontier of Artificial Intelligence (AI). In this context, Ehrlenspiel and Meerkamm~\cite{EhrlenspielMeerkamm17} see the greatest opportunities for influencing a design in the early phase of the development process. 

Design in context of this paper is an engineering activity, which uses scientific and engineering knowledge to find solutions for technical problems. Solutions are usually embodied in artifacts. Design is a very important activity which affects a lot of areas of human life and builds upon science and experience~\cite{Pahl.2007}.

Early on in the design process functional structures need to be found. These serve as the foundation in order to search for principal solutions before the artefact is structured into realizable modules. These functions represent demands of customers which they pose on the artefact.~\cite{VDI2019}

A function in the context of design describes a state or state transition of entities using operators. Storing a fluid or transforming kinetic energy into electric energy are typical examples of functions in this sense. When looking at a combustion engine, the overall function is to convert chemical energy into kinetic energy. In order to fulfill this function, several subfunctions like  guiding the fuel to the combustion chamber or converting linear kinetic energy into rotary kinetic energy are needed. The relations of the subfunctions can be mapped to a functional structure which represents the overall function. In general, not all subfunction and their relations to each other are known. It is therefore important to decompose the overall function into a functional structure. 

The design domain shares several similarities with the planning domain, e.g. functions can be seen as actions and entities share properties with planning states in the planning domain. The task in the planning domain is usually to manipulate states using actions. An example from that domain is transporting goods from one city to another using e.g. an airplane. Here a plane needs to fly to the starting city first, where it picks up the goods before flying to the destination city where the goods are disembarked. The flights and pick up actions may be seen as functions, while goods being at some place may be interpreted as entity states. These similarities can be of great use when solving function related problems in the design domain.

However, the decomposition task is not well automated and involves experienced design engineers to build correct functional structures for design problems. This leads to the following research questions (RQ).

\smallskip


\noindent \textit{RQ 1:} Can general Function Decomposition (FD) be automated in the same way, the planning domain handles its problems?

\smallskip

\noindent \textit{RQ 2:} What set of components is needed to realize an automated FD using tools from the planning domain?

\smallskip

\noindent \textit{RQ 3:} What algorithms, respectively tools are capable of realizing an automated solution to a FD problem?

\smallskip


The main contribution of this paper is that a functional decomposition problem (FDP) can be formalized as a planning problem due to its similarities. The task of finding a FD can then be automated by using an established planning algorithm.


The remainder of the paper consists of a state of the art section followed by the proposed approach, which is explained in Section \ref{sec:solution}. Two different application examples are then discussed and the results are being evaluated before the paper ends with a conclusion and outlook section.

\section{State of the Art}

In order to tackle the research questions above, the state of the art needs to be elaborated. The following subsections will consider the early design process, the definition of functions and their modelling. Afterwards the state of the art in regards to FD is displayed from both the Design and AI point of view. The final subsection clarifies the existing research gap.

\subsection{Early stage design process}

In this subsection, the early phase of the product development process is discussed in order to classify the concept of function. 
The guideline VDI 2221~\cite{VDI2019} deals with fundamentals of methodical development of technical products and systems. The guideline defines central objectives, activities and work results in a model of product development, which represent central guidelines for interdisciplinary application in industrial practice. 

Product planning and the release of the development order form a first step. From the definition of the requirements and the wishes, an initial list of requirements is provided as the base for conceptual design. An essential element of the design order is the overall function with the purpose of the product. Identifying functions and their structures is about clarifying what purpose the product, or a component of it, is intended to serve. Relevant functions and their structures are then determined and solution principles and their structures are searched for until a breakdown into feasible modules and their design can be made. After the individual modules have been combined into a complete product, the design and usage specifications are worked out. \cite{Pahl.2007}

Due to the parallelization of workflows, the interdependencies of mechatronic systems and the increasing reuse of solutions, it is becoming increasingly difficult to clearly delineate the individual phases. Design includes the part of conceptualization in which a basic idea of something is developed \cite{EhrlenspielMeerkamm17}. 

The development of this basic idea is made possible by the elaboration and representation of functions, function structures and their effects and effect carriers, their outline and effect structure.
The concept represents the first solution principle for a procedure. According to Ehrlenspiel and Meerkamm~\cite{EhrlenspielMeerkamm17}, a concept is generally understood as a "first version, plan, program for a procedure". Here, the solution principle is in the foreground. The term concept is also used for other specifications such as design, production, assembly, and sales concepts. Accordingly, designing includes organizational modeling in addition to technical modeling.

The early concept phase is of high economic relevance due to the cost influence. The functions represent the link between the basic idea, requirements, physical effects and the derivation of solution principles. 

These are therefore analyzed in more detail in the next section.

\subsection{Function definition}

In this subsection, the question of the definition and understanding of functions will be explored. The term function is used in various fields, e.g. in mathematics, work science, engineering or computer science. This article relates to the development of mechatronic systems, so a selection of definitions and modeling from this area is listed below.

According to Pahl/Beitz~\cite{Pahl.2007}, a function specifies the general and intentional relationship between the input and output of a system with the goal of performing a task. Thus, the utility of a system is described. Functions are not limited to technology and its physical and mathematical relationships. In particular, functions are considered according to value, impression, aesthetics or symbolism. Especially for mechanical systems, the respective transformations between the input and output flows are used for the analysis when considering the functions and are subdivided into material flows, energy flows and signal flows.

According to Koller~\cite{Koller:96340}, a technical function is a cause-effect relationship between an input and an output variable. This can be specified concretely by the description of the property or the state of the input and output variable as well as the activity, which is necessary for this conversion. 

Parallels can be drawn to the VDI/VDE 3682~\cite{VereinDeutscherIngenieureVerbandderElektrotechnikElektronikInfomartionstechnik.2005} which formalises process descriptions. Here processes are described by networks, where states are defined as nodes which are then linked to process operators. The result is a transfer from what the VDI/VDE 3682 calls a status \textit{ante} to a status \textit{post}.

Gero~\cite{Gero_1990} sees function as a link between the intended description of a design purpose and the physical behavior or structure of an artifact. The design follows the function and the created artifacts fulfill functions.

A state-based view on functions is taken by Roth~\cite{Roth.2000}, in which the changes of the three entities material, energy and information define general functions. According to Roth these entities are transferred from one state to another by the use of functions. This, as well, is close to the VDI/VDE 3682. Entities appear as the input and output states which are connected by a process operator i.e. the function itself. E.g. a stored liquid is guided to a certain location for further processing. The input entity is the material \textit{liquid} and the status is \textit{stored}. \textit{Guiding} is the function, respectively the process operator. The output status is \textit{guided liquid}.

Another parallel can be drawn to the Planning Domain Definition Language (PDDL). Here so called \textit{actions} transfer preconditional states into effect states~\cite{Russell.2021}. Where the former can be seen as the input state and the latter can be seen as the output state.

This allows all machine systems to be described in terms of their functional structure, so that it can be used to develop new machines with similar purposes. Target functions only specify the desired relationships, but not their physical or chemical realization. Accordingly, a distinction is made between general and specific functions.

Deng~\cite{Deng.2002} describes functions as intended behavior. Functions are performed in a flow. The desired behavior of a product is hence implemented by a function. He distinguishes between purpose functions and action functions where the former aim at the overall function of the product and are realized by the ladder. These action functions can be described in terms of the Roth approach.

In this section, different views of the concept of function were taken. It can be seen that these different concepts can lead to a different mapping logic and can have an influence on the FD.

\subsection{Functional modeling}

In the subsection before, the definition of functions was discussed. Now the question of how functions are modeled shall be explored. The modeling of the transformation processes for entities are the common basis for an integrated functional modeling.

Erden et al.~\cite{Erden2008} reviewed function modeling approaches and applications and proposed a general framework for harmonizing existing approaches. One recommendation is to separate the subjective and objective domains in terms of design objects and to use functions as a subjective category to link these two domains. 

The paper by Srinivasan et al.~\cite{srinivasan2012framework} presents a framework for describing functions in the design process. For this purpose, a comprehensive review is performed on representations of functions and the views considered there. From this point, a framework is derived, which represents abstraction levels, requirements/solutions and the system/environment from the user's/developer's point of view. Not all functions can be made explicit, because some human aspects are not yet formalizable.

The framework includes a requirements-solution model and a system-environment model, based on which the functions can be uniquely assigned in a dedicated manner. An important point is the distinction between deducible and possible functions. Deducible functions can be assigned to known solution principles. Possible functions consider an implicit ability to derive potentially new solution principles or to use existing solutions for further sub-functions. 

Another circumstance that can be pointed out by the framework is the creation of variants in the context of FD. However, this is not directly referred to, because the mapping locality requires an extension of the product model.~\cite{Demke2021}. 

Chakrabarti and Blessing~\cite{chakrabarti_blessing_1996} describe three different formal representations of functions with the context of supporting activities on computers. These are verb-noun pairs, input-output flows and transmissions between input-output situations. Verb-noun pairs like "transmit torque" involve the mapping of functions in natural language, which is the most common form. 

The input-output flows of energy, material or information refers to Weizsacker, Rodenacker, Koller and Roth. According to Roth~\cite{Roth.2000} there are 30 normalized functions which correlate with all technical products. 

\begin{figure*} [t]
    \centering
    \includegraphics [width=0.8\textwidth]{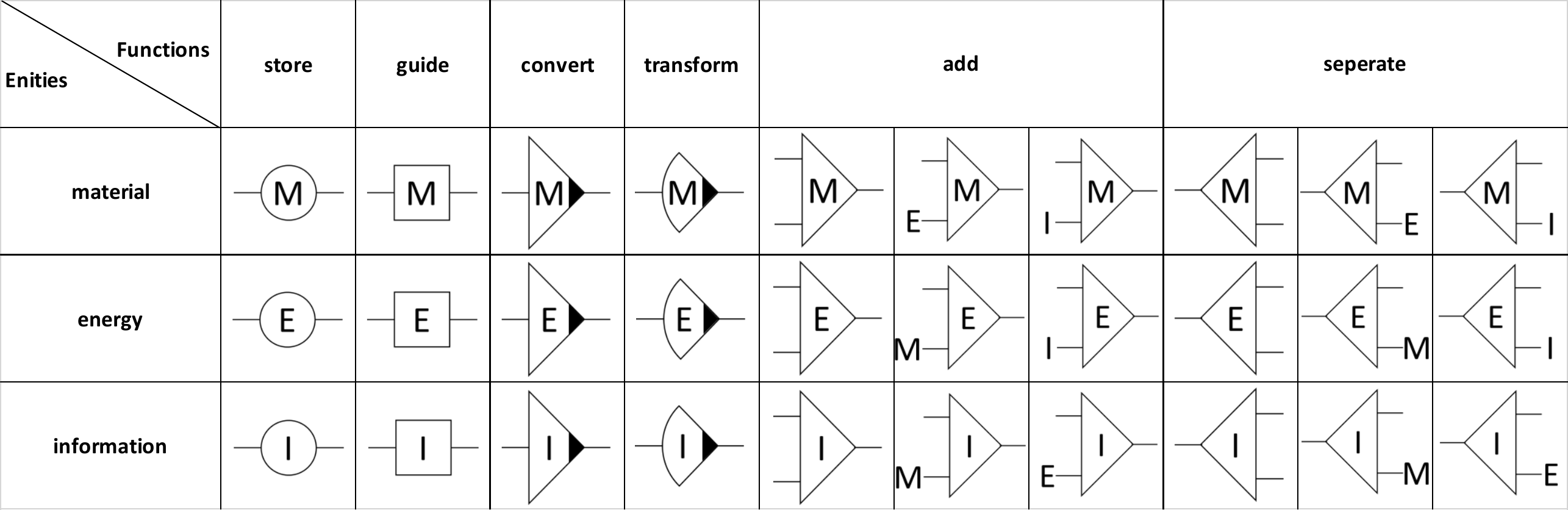}
    \caption{30 functions according to Roth\protect\cite{Roth.2000}}
    \label{fig:Roth}
\end{figure*}

These functions link general entities, i.e. material, energy and information with general operations. These operations are storing, guiding, transforming
, converting
, summative and distributive linking. Where each of the ladder two operations can be further divided considering whether equal or unequal entities are linked.
The transformation between input-output situations is based on Hubka~\cite{Hubka.1984}. An overview of Roth's function is given in Figure \ref{fig:Roth}.

Eisenbart et al.~\cite{eisenbart_gericke_blessing_2013} collaborated on an analysis of cross-disciplinary functional modeling. They analyze 41 different approaches and categorize them. The perspectives are states, effects, transformations processes, interaction processes, use cases, technical system allocation and stakeholder allocation. Therefore, they show that there is no such shared sequence for functional modeling across disciplines. In a further analysis, they work out, that modeling the transformation process provides a common basis for modeling across disciplines. Special views from the disciplines are to be linked with the transformation perspective. Individual character inventories can be used in the process. 

Dybov and Stark~\cite{DYBOV2021391} show that function models are considered with respect to their transfer between mechatronic products. With the representation of functional architecture levels, they show a way to map highly networked mechatronic production. It is essential that criteria are presented to derive a common functional model between at least two products. The determination of the structure of the FD and the function flow are in the center. They find that all functions and subfunctions from which the common function model is derived have the same input and output signals and the same semantic expression. 

Different modeling approaches for functions were considered. Functions can be described by natural language, mathematical expressions or diagrams, graphs or geometry. For engineering purposes, verb + noun pair descriptions are particularly suitable, since they can be easily used in the design process. 

Limitations arise from a non-uniform description. Roth offers a comprehensive library of functions for the design of mechanical products. The use of general functions is important here, as these form the base for specific models. Having covered the definitions and models, a general commonality is that functions can be understood as a transformation process in which states of entities can be transferred from one to another.

The next subsections will be a discussion of existing approaches to FD in engineering as well as in computer science respective AI science. 


\subsection{Design approaches in Function Decomposition}

In this subsection, approaches from the field of product design research are listed. 

In general, the term FD can also be found under the terms decomposition, logical synthesis, abduction and disassembly. As already mentioned, the overall function of a product is not directly solvable and is also subject to uncertainties. The FD describes the decomposition of the overall function into manageable subfunctions, or subproblems~\cite{Pahl.2007}.

The decomposition of the overall function into subfunctions is a basic procedural strategy for solving complex problems in methodical product development. At the beginning, a detailed analysis of the overall function is performed before a suitable subdivision into subfunctions is carried out by decomposition. The partial functions can be traced back to elementary functions. The solution is the function structure. This is accompanied by an analysis of the interdependencies at the respective levels of detail. 

The goal of this function structure synthesis or function synthesis is to show possible functional structures and indicating the most favorable one.

The synthesizing of the function structure is done by logical linking of defined partial or elementary functions, so that the required cause-effect relationships of the overall system become visible~\cite{Koller:96340}. It breaks down the synthesis of function into the following steps. First describing the purpose respectively main or core function, then structuring and developing one or more corresponding sub-functional structures and selecting the most favorable structure and finally breaking down into alternative basic operation structures and select the basic operation structure that appears to be the most favorable.

Further work e.g. by Birkhofer~\cite{Birkhofer} and Matthiesen~\cite{Matthiesen} shows the interactions of functions to their effect models and principle solutions.

Roozenberg~\cite{Roozenburg.2002} demonstrates that two fundamentally different forms of abduction can be distinguished - explanatory abduction and innovative abduction. What is usually understood by ``abduction'' is explanatory abduction, but synthesis in the sense of reasoning from function to form follows the pattern of innovative abduction, or ``innoduction'' This means that the form does not indisputably follow from the functions to be fulfilled and that, in principle, there are always a lot of feasible solutions. The reasoning from function to form is a creative process, which can be encouraged methodically, but cannot be logically guaranteed.

Research from the design perspective is focused on methodologies for the decomposition process to guide engineers but does not answer the question of how the functional decomposition task could be automated.


\subsection{AI approaches in Function Decomposition}

In this subsection, approaches from the field of artificial intelligence research are listed. 

A computational knowledge based approach to FD is shown by Umeda et al.~\cite{umeda_ishii_yoshioka_shimomura_tomiyama_1996}. They show a formalism about representation and reasoning to support semantic and physics-based reasoning on the information hidden in the plain-english flow terms. The function-behavior-state diagram is a base for task and causal decomposition. In addition to providing elements for function modeling and FD, it also provides a knowledge base. The approach aims to support the design engineer fulfilling the task rather than automating it.

Yuan et al.~\cite{Yuan.2017} describes an approach to automated functional decomposition based on morphological changes in material flows. The approach refers to three decomposition principles, namely the principle of material conservation, the conservation of form elements and the energy minimization principle. Material conservation means that the sum of all input flows of a material flow must be equal to the output flows. Accordingly, the principle can be used to plausibilize input and output during decomposition. The conservation of shape elements says that the total shape change of a material flow is based on the local relationship changes between all shape elements. The principle of energy minimization states that the best solution for the morphological changes of a material flow is when the energy consumption is minimal. The approach is only concerned with changes in flows for the entity material. It lacks a generalization for the entities energy and information.

Tensa et al.~\cite{Tensa2019} shows an approach on how to learn association rules between functions, parts and flows via an a priori algorithm. This approach builds on and analyzes existing product data. The corresponding function structures can be derived from the object dependencies and thus support the decomposition process. 

In the paper of Feldman et al.~\cite{Feldman.2019}, the creation of circuit designs are automated via the use of a Quantified Boolean Formulas (QBF) Solver by reducing the design generation and testing with new algorithms to the QBF format. In order to achieve this, they propose new algorithms for design and design space exploration. The designs computed by these algorithms are compositions of function types specified in component libraries. The algorithms they present are sound and complete and are guaranteed to discover correct designs of optimal size, if they exist. They apply a method to the design of Boolean systems and discover new and more optimal classical and quantum circuits for common arithmetic functions such as addition and multiplication. 

In Mao's~\cite{Mao2019} work, designers should be able to apply function-based design methods even without appropriate knowledge. For this purpose, a formalism of representation and inference is proposed to enable semantic and physical inference from the hidden information in plain-text flow labels. 
This is intended to enable automated decomposition of black-box functional models and generation of multiple design alternatives. The use of plain-text flow relationships avoids the need for standardized libraries such as those presented by Roth~\cite{Roth.2000}, which is usually the case with graph-based function modeling. The definitions of the function classes in the library in combination with chosen grammars, which are defined on different formalism levels, represent thereby the basis for the reasoning and further the function decomposition.
Semantic inference is used to derive the changes in flow types, flow attributes, and the direction of these changes between the input and output flows associated with the black box. 
Qualitative physics representation is then used to determine the material and energy exchanges between the flows and the functional properties required for these exchanges.
The topological layer allows arguments to derive several different functional features in topologies, resulting in several alternative decompositions of the functional black box. 
The flow phases, flow attributes, qualitative value scales for the properties, and qualitative physical laws are formalized via the data representation. A three-layer algorithm manipulates this data for inference. 

\subsection{Discussion}

The decomposition process has far-reaching interdependencies due to the adjacent development steps, such as the clarification of requirements or the derivation of effect structures and the search for principle solutions. According to this interdependency, there are several decomposition directions, which lead to the fact that several decomposition principles have to be used at the same time. 

This section has shown, that functions by their definition and their modelling share similarities with alien domains like process management and the planning domain. While research in design is mainly focused on the approach to understand and improve the methodology of functional decomposition to support the design engineer in the design process, AI research has interest in the decomposition task as an application for AI based methods. The presented approaches either aim only to support the design engineer in the task or just handle  the automation of specific decomposition tasks which cannot easily be generalized for general functional decomposition problems.

To the knowledge of the authors there is no approach existent yet, which takes advantage of the aforementioned similarities between functions and planning states to solve decomposition problems using well established algorithms from the planning domain.

\section{Solution}\label{sec:solution}

The core idea of this paper is to interpret function decompositions---i.e. the design problem---as a planning problem: Functions in design may be formalized as actions in planning algorithms. A function decomposition is then correct if and only if the plan leads to the desired goal. 

The advantage is that in the planning domain a number of well known and rather efficient algorithms exist to solve the planning problem. By mapping FD to a planning problem, these algorithms can be used to compute efficiently a decomposition. This plan may then be represented as a graph, i.e. a functional structure, to further human understanding.

In connection to the discussion above and to answer the resulting research questions, necessary formalisms need to be introduced first.

\subsection{Planning Problem and PDDL} \label{subsec:problem}

\textit{PDDL} is a standardized well established language for AI planning problems which comes with several efficient solution implementations. The language describes a problem in First Order Logic (FOL) by giving the initial state, stating the goals which solve the problem, naming all possible actions available in a state and stating the result of executing an action.

\medskip
\noindent A planning problem (PP) is a tuple $(S,i,g,A)$: 

\noindent $\bullet$ $S$ is the set of states where $s\in S, s= (b_1 \land ...\land b_j), j \in \mathbb{N}$ and $b_i$ is an atomic symbol in the logic calculus.

\noindent $\bullet$ $i \in S$ is the initial state, i.e. the situation at production start.

\noindent $\bullet$ $g \in S$ is the goal state, i.e. the situation at production end.

\noindent $\bullet$ The set $A$ is the set of all actions. An action $a\in A$ is a tuple $(n,p,e)$. The name $n$ is a unique string which denotes the action. The precondition $p$ is a conjunction of atoms or negation of atoms. The actions can be executed iff the system is in a state $s$ with $s \models p$. The effect $e$ is also a conjunction of atoms or negation of atoms. A resulting state $s'$ of executing action $a$ from state $s$ is defined as $s \land e$.





The solution to a planning problem is a plan $p=[a_1,a_2,...,a_{k-1},a_k], a_i\in A$, which is a finite sequence of actions. By executing these actions in the order of the plan a goal state $s'\models g$ is reached from the initial state $i$. 


\subsection{Decomposition Problems and Roth Functions}\label{subsec:decription}

While the idea from above---using planning---solves our algorithmic problem, one problem remains: Which are the basic functions (i.e. the actions)? 

Roth~\cite{Roth.2000} provides a solution by defining a general set of functions, expressed as with verb + noun pairs. He defined general functions which---combined with the three entities material, energy and information---add up to thirty functions which comprise all functional aspects of products. 

As shown in Figure \ref{fig:Roth} all functions are represented by a symbol. Each symbol has inputs to the left of it and outputs to the right of it. Any product can be functionally decomposed into these fundamental functions. This holds not only true for mechanical or electronic products but also for software.

\medskip
\noindent A FDP on behalf of Roth is a tuple $(S_{ae},i_d,o_d,F)$: 

\noindent $\bullet$ $S_{ae}$ is the finite set of allocated entity states of the FDP. An allocated entity state $s_{ae} \in S_{ae}$ is defined as a conjunction of allocated entities in the form:
$$
s_{ae} = (ae_1 \land ...\land ae_i);\, i \in \mathbb{N}
$$ 

\noindent $\bullet$ The allocated entities are directly derived from the Roth functions. In addition to these, an entity may also be absent. An allocated entity is defined as:
\begin{eqnarray*}
ae \in \{Stored(material),..., \\ 
SeparatedEnergy(information) \\
\neg (material), \neg (energy), \\
\neg (information)\}
\end{eqnarray*}

\noindent $\bullet$ The input to the FDP $i_d \in S_{ae}$ denotes to the given allocated entities. 

\noindent $\bullet$ In the same way $o_d \in S_{ae}$ denotes to the desired output  in the form of allocated entities.

\noindent $\bullet$ The set $F$ is the set of all Roth functions. A function $f \in F$ is a tuple $(n_f,i_f,o_f)$. The name $n_f$ is a unique string which denotes the function. The input $i_f$ is a conjunction of allocated entities. Each state $s_{ae} \models i_f$ is a state in which the function $f$ may be executed. The output $o_f$ is also a conjunction of allocated entities. A resulting state $s'_{ae}$ of executing function $f$ in the state $s_{ae}$ is defined by removing the input of $f$ from state $s_{ae}$ and adding the output of $f$ to $s_{ae}$.




It is very important to point out the similarity between this representation and the representation of actions in a planning problem. 
The solution to a FDP is a functional decomposition $fd = [ae_1, ae_2,...,ae_{k-1},ae_k]; \, k\in \mathbb{N}$, which is a finite sequence of functions. By executing these functions in the order of the $fd$, a desired state $s'_{ae} \models o_f$ is reached from the given input $i_f$.


\subsection{Mapping Decomposition Problems onto Planning Problems}\label{subsec:map}

We can now map a function decomposition problem FDP = $(S_{ae},i_d,o_d,F)$ (see section \ref{subsec:decription}) onto a planning problem PP = $(S,i,g,A)$ (see section \ref{subsec:problem}).

Given FDP = $(S_{ae},i_d,o_d,F)$  we create PP = $(S,i,g,A)$ in the following steps:
\begin{description}
\item[$S$:] $S$ is computed as by creating all possible allocated entity states. 
\item[$i$:] $i$ is computed as by creating the initial state as a conjunction of the input allocated entities.
\item[$g$:] $g$ is computed as by creating the goal state as a conjunction of the output allocated entities.
\item[$A$:] $A$ is computed as by creating actions from all Roth functions. 
\end{description}

\noindent This as well is in line with the description of a planning problem in PDDL. By replacing the expressions \textit{input} and \textit{output} with \textit{precondition} and \textit{effect}, all Roth functions can be represented in PDDL.



\subsection{Planning Algorithm}

The previous subsection has shown that a decomposition problem can be represented as a planning problem, i.e. in PDDL. Solving planning problems represented in PDDL is a well known task in the AI domain. Several algorithms for this task are already known to work well. By stating a decomposition problem in the same way a planning problem is stated, the planning algorithms should be able to find a functional structure for a given decomposition problem analogous to finding a plan. Since the allowed actions, in the case of a decomposition problem, the Roth functions, do not change from problem to problem the only difference between the decomposition problems is the input and desired output respective the initial state and goal state. 

This directly answers \textit{RQ 1}. By representing a decomposition problem in the way shown above, it becomes similar to a planning problem stated in PDDL and is thus solvable by algorithms used in the planning domain. 

It also gives an answer to \textit{RQ 2}. The components needed to automate the decomposition are the Roth functions represented in PDDL, as well as the desired output of the overall product function and as well as its inputs. Both inputs and outputs need to be stated as respective conjunctions of allocated entities.

The solver used for this approach is the \textit{Partial Order Plan} algorithm (POP). The partial order approach is suitable when subproblems may occur independent from one another and it works well on decomposable problems. POPs are often used where understanding the plan by a human operator is important.~\cite{Russell.2021}

The specific POP algorithm used here is an adaptation from Russell and Norvig~\cite{Russell.2021}. Its mode of operation applied to decomposition problems is shown in Algorithm \ref{alg:Partial}. 

In line 1 of the algorithm the $Agenda$ is initially filled with the ordered pair $(G, finish)$. Here $G$ is the intended output $o_d$ and $finish$ acts as a helping function which has $G$ as input $i_f$. $Agenda$ itself is a set of ordered pairs. The set $Functions$ in line 2 is initially filled with the functions $start$ and $finish$. Analogue to $finish$, $start$ is a function which has $i_d$ as output $o_f$. The set $Constraints$ stores initially the constraint, that $start$ needs to be executed before $finish$. $CausalLinks$ is initially an empty set. As long as the $Agenda$ is not empty the following steps will be executed. In line 6 a pair of input $G$ and its function $fct_1$ is picked from the $Agenda$. A function $fct_0$ which has $G$ as output is then chosen from the Roth functions in line 7. Then in line 8 the pair $(G, fct_1)$ is removed from the $Agenda$. In line 9 $fct_0$ is added to the set of $Functions$. In line 10 the constraint, that $fct_0$ needs to be executed after $start$ is added. From line 11 to 13 the algorithm checks if $fct_0$ is a threat to the existing constraints and if so, it tries to sort $fct_0$ prior or posterior to the threatened constraint. The set $(P, fct_0)$ is then added to $Agenda$, with $P$ being the input $i_f$ of $fct_0$ in line 14. The certainty that $fct_0$ is executed before $fct_1$ is added to $Constraints$ in the following line. In Line 16 the set $CausalLinks$ gets updated with the ordered triple $(fct_0. G. fct_1)$. Here $G$ links $fct_0$ with $fct_1$ by being output of $fct_0$ and input of $fct_1$. From line 17 to line 19 the algorithm checks if any function in $Functions$ threatens the just described ordered triple and if so tries to sort the concerning function prior or posterior. In line 21 the algorithm outputs the ordering functions, if possible, given $i_d$ and $o_d$.

\begin{algorithm} [t]
    \caption{Function Decomposition Solver}
    \label{alg:Partial}
    \textbf{Input:} Goals: set of entity states to achieve\\
    \textbf{Output:} Function Decomposition
    \begin{algorithmic}[1] 
        \State $Agenda \gets \{(G,finish):G \in Goals\}$
        \State $Functions \gets \{start,finish\}$
        \State $Constraints \gets \{start<finish\}$
        \State $Causal Links \gets \{0\}$
        
        \While{$Agenda \neq\{\}$}
            \State select $(G, fct_1)$ from $Agenda$
            \State choose $fct_0$ that achieves $G$
            \State remove $(G, fct_1)$ from $Agenda$
            \State $Functions \gets Functions \cup \{(fct_0)\}$
            \State $Constraints \gets add\_constraint \{$ \par $start < fct_0,Constraints\}$
            
            \For{each $CL \in Causal Links$}
                \State $Constraints \gets protect($ \par \quad $\>$ $Cl,fct_0,Constraints)$
            \EndFor
            
            \State $Agenda \gets Agenda\cup\{$ \par $(P,fct_0)$: $P$ is input of $fct_0\}$
            \State $Constraints \gets add\_constraint$\{ \par $fct_0<fct_1, Constraints\}$
            \State $Causal Links \cup \{(fct_0,G,fct_1)\}$
            
            \For{$each\, F \in Functions$}
                \State $Constraints \gets protect$\{ \par \quad $\>$ $(fct_0,G,fct_1), F, Constraints\}$
            \EndFor
            
        \EndWhile\\
        
        \Return total ordering of Functions
    \end{algorithmic}
\end{algorithm}

The described Function Decomposition Solver gives an answer to \textit{RQ 3}. A modified POP algorithm is capable of automatically solving decomposition problems, given the problem is stated in PDDL and a decomposition exists. If, in addition, the number of iterations is limited, the decomposition must be found in a feasible amount of time.

\section{Empirical Results}

In this section the presented approach is put into practice and evaluated using the examples of a coffeemaker and a siege engine.

\subsection{Coffee Maker}

The first example deals with a device which is supposed to make coffee based on a few given constraints. The device has initially access to stored water and to electric energy. In addition, a potential user might find it convenient if coffee powder is directly stored in the device. These three allocated entities serve as input into the overall function of the device. The overall goal, the device shall achieve, is to make hot coffee. So the overall function output is hot coffee which can generally be described as water loaded with thermal energy, which is added to coffee powder. In line with Subsection \ref{subsec:decription} the decomposition problem can be represented as:

\smallskip

\begin{flushleft}

\noindent $i_d = (Stored(water) \land Stored(electric) \land $ \par $ Stored(powder))$ \newline
\noindent $o_d = (Converted(electric) \land AddEnergy(water) \land AddMaterial(powder))$ \newline
\noindent $F =(RothFunction1),..., (RothFunction30)$

\end{flushleft}

\smallskip

\noindent The previously described Function Decomposition Solver returns the following output when presented with the problem:

\smallskip

\begin{flushleft}

$[\{start\},$ \newline
$\{GuideMaterial(water),$  
$ GuideEnergy(electric)\},$ 
$ \{ConvertEnergy(electric)\},$ 
$ \{GuideMaterial(powder), AddeM(electric, water)\},$ 
$\{AddmM(water, powder\},$ \newline
$ \{finish\}] $

\end{flushleft}

\smallskip

\begin{figure} [t]
    \centering
    \includegraphics [width=0.9\columnwidth]{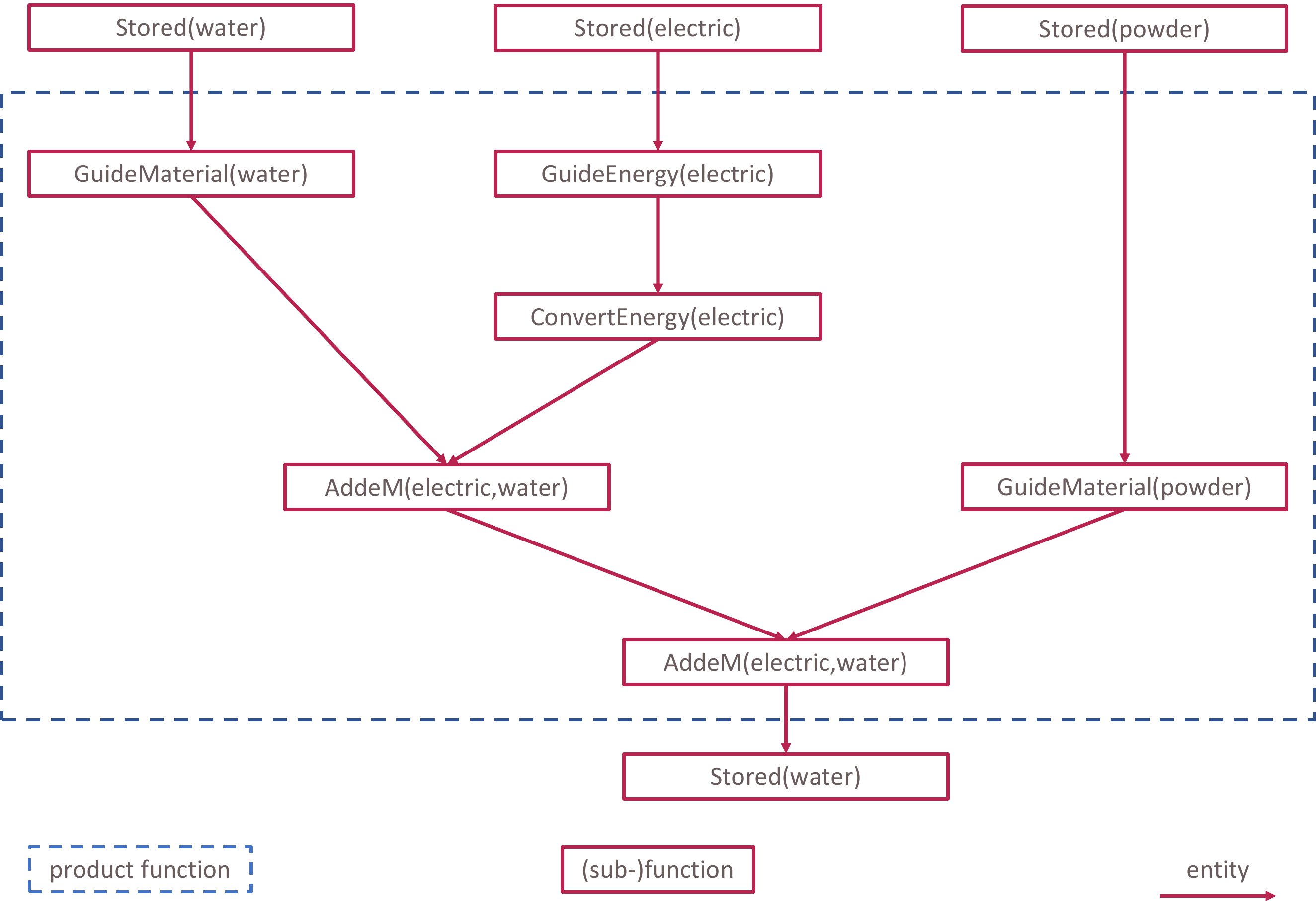}
    \caption{derived functional structure of the coffee maker device}
    \label{fig:CoffeeMaker}
\end{figure}

The order in which the functions occur is given by the curly brackets. It does not matter in which order the functions are executed inside these curly brackets. E.g. it does not make any difference if the water is guided first or if the electricity is guided first or if both function occur simultaneously. These subfunctions occur independent from each other. It does however matter, that the electricity is first guided before it is converted. The solver output states, that first water and electricity need to be guided, before the electricity is converted. Then the coffee powder has to be guided and the converted energy needs to be applied to the water. Finally the water loaded with the converted energy is added to the powder, which results in the desired output of the overall function of the device. Figure \ref{fig:CoffeeMaker} shows the plan as a graph.

The Function Decomposition Solver was able to automatically decompose the overall function given the required inputs by using the Roth functions translated into PDDL and applying the modified POP algorithm to the problem.


\subsection{Siege Engine}

The second example is a device which has the purpose to destroy a solid wall by applying force to it. Initially available are only stored human power and a solid material like e.g. some kind of timber, which is also stored. The idea is to guide the timber loaded with sufficient energy into the wall, in order to collapse it. As it turns out, the wall is quite resilient and the used power needs to be converted first. By representing these contraints and the Roth functions in PDDL, the problem can be stated in the following way:

\smallskip

\begin{flushleft}

\noindent $i_d = {Input}(Stored(kinetic) \land Stored(timber))$ \newline
\noindent $o_d = (Transformed(kinetic) \land AddEnergy(timber))$
\noindent $F = (RothFunction1),..., (RothFunction30)$

\end{flushleft}

\smallskip

\noindent The Function Decomposition Solver returns the following solution:

\smallskip

\begin{flushleft}

$[\{start\},$ \newline
$\{GuideEnergy(kinetic),$  
$ GuideMaterial(timber)\},$ 
$ \{TransformEnergy(kinetic)\},$ 
$ \{AddeM(kinetic, timber)\},$ \newline
$ \{finish\}] $

\end{flushleft}

\smallskip

\begin{figure} [t]
    \centering
    \includegraphics [width=0.9\columnwidth]{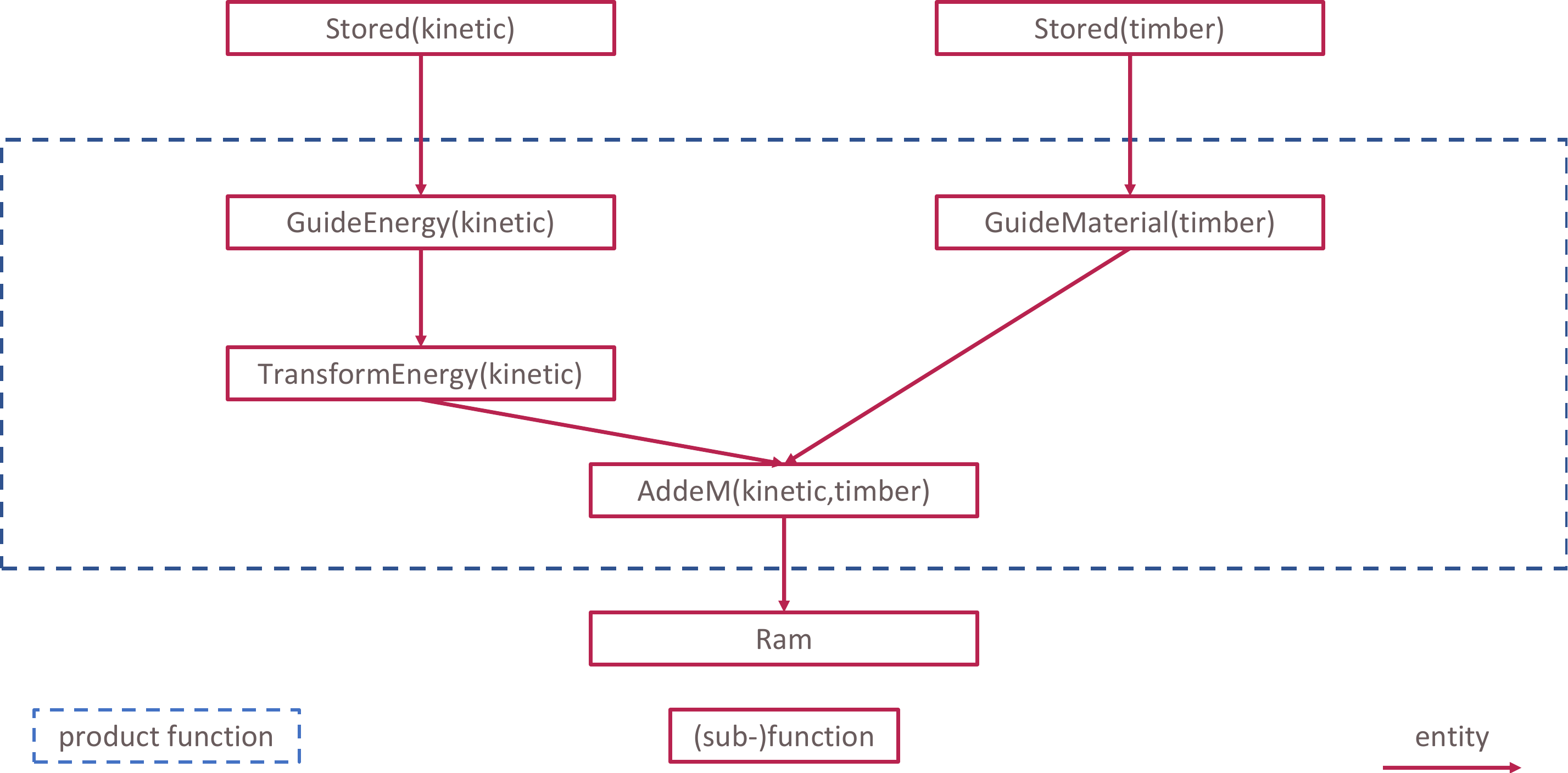}
    \caption{derived functional structure of the siege engine device}
    \label{fig:SiegeEngine}
\end{figure}

Similar to the first example, the order in which the kinetic energy and the timber material are being guided does not matter for the overall function as long as this happens prior to the energy transformation. One could argue here, that even the guidance of the timber may take place after the energy is being transformed, which will also lead to the device fulfilling its overall function. While this is true, the solution the solver has to offer is also correct. The solution for this siege engine problem solved by the algorithm is depicted in Figure \ref{fig:SiegeEngine}.

Here as well the solver used the given inputs and desired outputs as well as the Roth function represented in PDDL to successfully generate a FD for the device by using the modified planning algorithm.


\subsection{Discussion}

The evaluation showed that an automated functional decomposition can be successfully performed in both application examples. The solution quality was evaluated according to the completeness of the functions and partial functions, as well as according to the logically correct sequence of the preceding rankings. The results were discussed and evaluated by experts from the engineering sciences.

\section{Conclusion and Outlook}


The work in this paper has shown, that in regards to \textit{RQ 1} it is possible to automate FD in the design domain by representing the problem in PDDL and use a well known planning algorithm to find a correct solution. This approach of handling a FD as a planning problem is, to the knowledge of the authors, new.


The presented approach makes use of the Roth functions, which are a known standard in the design community. These functions can be represented by a verb + noun pair which can be represented in PDDL. In addition to these functions, the overall input and output of the to be designed product function can be stated similar to initial states and goal states known from the planning domain. This answers \textit{RQ 2}.


Due to the representation in PDDL, it becomes possible to solve decomposition problems using well known planning algorithms. The authors opted for an adaptation of the POP algorithm to answer \textit{RQ 3}, since it works well with independent subproblems and the output is well understandable for human operators. It has been shown that the presented approach works by providing two examples.


External interdependencies from requirements, physical effects, effect models, and principle solutions were not directly considered, because there is a high complexity in design domain and pre-processing steps that need to be prepared. Only general functions and selected requirements were used, so that the specification is omitted. A further point refers to the formation and the modeling of variants of products, which were not sufficiently considered.

Preparatory work can be derived from the 30 Roth functions themselves. Requirements lists need to be written in the verb + noun standard given by these functions. It has been shown that from these requirements a meaningful general functional structure can be derived. If a requirements list is created disregarding this convention a translation into this standard needs to be performed by an operator.

Regarding the kinds of requirements used in the design domain, non-functional requirements and insufficiently formalized requirements were not considered further, so that further development would be preprocessing. Another point arises from the decomposition principle of planning and the possible combinations of functions. The formulation of planning problems requires that initial states and goal states as well as reference functions are named. This requires further preprocessing steps, which are not implemented in this paper, because the core is the general decomposition.

Further research on functional decomposition can be done on several areas based on this paper. One point is the application of this approach in the context of assistance systems for the early phase of the product development process. Another point concerns the inclusion of further interdependencies such as the requirements list, physical effects, effect models, principle solutions and design solutions. As mentioned works show, there are interdependencies that affect decomposition. The question of how are these dealt with in terms of a ranking problem arises. 

Further, the formalizability of information related to humans was also pointed out. Again, the interactions between human and assistance in decomposition can lead to further insights. This can possibly also be relevant for creative design research by extending empirical findings through AI based approaches.

From a computer science point of view, further research may be done concerning the used algorithm to solve the problem. The question arises what algorithms are also suited to find solutions for the elaborated approach. A rational next step would be to rank these algorithms based on their performance and solution quality to solve functional decomposition problems of varying complexity. Further research may be invested in the ability to adapt to different decomposition problems. For instance, is it possible, to adjust the approach and algorithm to also solve more specialized decomposition problems like \textit{Intensity and Quantity} functional structures?

\printbibliography

\end{document}